\documentclass[a4paper,10pt]{scrartcl}
\usepackage[left=2cm, right=3cm, top=2cm]{geometry}
\usepackage[english]{babel}
\usepackage{amsmath}
\usepackage[utf8]{inputenc}
\usepackage{graphicx}
\usepackage{url,hyperref,lineno,microtype,subcaption,tabularx}

\title{A Neural Spiking Approach Compared to Deep Feedforward Networks on Stepwise Pixel Erasement}

\author{René Larisch\,$^{1}$, Michael Teichmann\,$^{1}$ and Fred H. Hamker\,$^{1}$}

\begin{document}
	\maketitle
	$^1$ \textit{Department of Computer Science, Chemnitz University of Technology, Chemnitz, Germany} 
	
	\begin{abstract}
		In real world scenarios, objects are often partially occluded.
		This requires a robustness for object recognition against these perturbations.
		Convolutional networks have shown good performances in classification tasks.
		The learned convolutional filters seem similar to receptive fields of simple cells found in the primary visual cortex.
		Alternatively, spiking neural networks are more biological plausible.
		We developed a two layer spiking network, trained on natural scenes with a biologically plausible learning rule.
		It is compared to two deep convolutional neural networks using a classification task of stepwise pixel erasement on MNIST.
		In comparison to these networks the spiking approach achieves good accuracy and robustness.
	\end{abstract}

	\section{Introduction}

	Deep convolutional neural networks (DCNN) have shown outstanding performances on different object recognition tasks \cite{Ciresan2012,Krizhevsjy2012,Simonyan2015}, like handwritten digits (MNIST \cite{LeCun1998}) or the ImageNet challenge \cite{Russakovsky2015}.
	Previous studies show that filters of a DCNN, trained on images, are similar to receptive fields of simple cells in the primary visual cortex of primates \cite{Cichy2016,Wen2017} and thus have been suggested, to a certain degree, as a model of human vision, despite the fact that the back-propagation algorithm, does not seem to be biological plausible \cite{Bengio2015,Diehl2015}.\\
	Alternatively, many models have been published in the field of computational neuroscience, whose unsupervised learning is based on the occurrence of the pre- and postsynaptic spikes.
	For example, Diehl and Cook presented a model using a spike-timing-dependent plasticity (STDP) rule to recognize digits of the MNIST dataset \cite{Diehl2015}.
	We propose a STDP network with biologically motivated STDP learning rules for the excitatory and inhibitory synapses to better mirror the structure in the visual cortex.
	We use the voltage based learning rule from Clopath et al. (2010) \cite{Clopath2010} for the excitatory synapses and the symmetric inhibitory learning rule from Vogels et al. (2011) \cite{Vogels2011} for the inhibitory synapses.
	During learning, we present natural scenes to the network \cite{Olshausen1996}.
	The thereby emerging receptive fields \cite{Clopath2010} are similar to those of simple cells in the primary visual cortex \cite{Hubel1962,Jones1987}.
	After learning ,we present digits of the MNIST data set to the network and measure the activity of the excitatory population.
	The measured activity vectors on the training set are then used to train a support vector machine (SVM) with a
	linear kernel to be used on the test set to estimate the accuracy of the neural network.\\
	We previously evaluated the robustness of classification by a gradual erasement of pixels in the MNIST data set.
	Our evaluation showed, that inhibition can improve the robustness by reducing redundant activities in the network \cite{Kermani2015}.
	To evaluate our spiking network, we apply this task by placing white pixels in $5\%$ steps in all images of the MNIST test set and by measuring the accuracy on these degraded digits.
	We compare our spiking network with two DCNNs.
	The first DCNN is the well known LeNet 5 network from Yan LeCun et al. (1998) \cite{LeCun1998}.
	The second one is based on the VGG16 network from Simonyan and Zisserman (2015) \cite{Simonyan2015}.
	Both deep networks are trained on the MNIST data set and the accuracy is measured on the test set with different levels of pixel erasement.\\
	We here follow the idea, that a biologically motivated model trained by Hebbian Learning on natural scence should discover a codebook of features that can be used for a large set of classification tasks. 
	Thus, we train our spiking model on small segments of natural scenes. 
	As these image patches contain different spatial orientations and frequencies, we obtain receptive fields which are selective for simple features. 
	With this generalized coding, we archived a recognition accuracy of $98.08 \%$.
	Further, our spiking network shows a good robustness against pixel erasement, even with only one layer of excitatory and inhibitory neurons.\\

\section{Methods}

Both deep convolutional networks are implemented in Keras v.2.0.6\cite{Chollet2015} with tensorflow v.1.2.1 and Python 3.6.
Our spiking network is implemented in Python 2.7 with the neuronal simulator ANNarchy (v.4.6) \cite{Vitay2015}.
To classify the activity vectors of our network, we used a support vector machine with a linear kernel, using the \textit{LinearSVC} package from the sklearn library v.0.19.1.

\subsection{Spiking model}

\subsubsection{Populations.}
\begin{figure}
	\centering
	\includegraphics[width=0.9\textwidth]{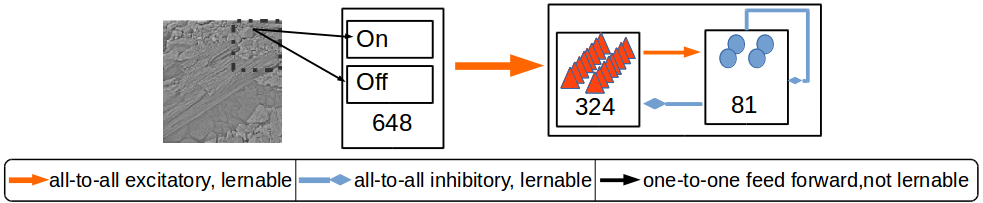}
	\caption{Schematic diagram of the spiking network. The input layer consists of 648 neurons. The second layer consists of 324 excitatory and 81 inhibitory neurons.}
	\label{fig_Net}
\end{figure}

The architecture of our spiking network (Fig.\ref{fig_Net}) is inspired by the primary visual cortex and consists of spiking neurons in two layers.
The input size is $18\times18$ pixels.
We used randomly chosen patches out of a set of whitened natural scenes to train the network.
To avoid negative firing rates, positive values of the patch are separated in an On-part and negative values in an Off-part.
Therefore, the first layer consists of 648 neurons in a $18\times18\times2$ grid.
Every pixel corresponds to one neuron in the layer.
The neurons fire according to a Poisson distribution, whose firing rate is determined by the corresponding pixel values.
The presented pixels are normalized with the absolute maximum value of the original image and multiplied with a maximum firing rate of $125 Hz$.
Each patch was presented for $125ms$. The learning was stopped after $400.000$ patches.
The presented patch was flipped around the vertical or horizontal axis with a probability of $50\%$ to avoid an orientation bias \cite{Clopath2010}.\\
The neurons in the first layer are all-to-all connected to the neurons in the second layer.
The second layer consists of a population of 324 excitatory and 81 inhibitory neurons to achieve the $4:1$ ratio between the number of excitatory and inhibitory neurons as found in the visual cortex \cite{Beaulieu1992,Potjans2014}.
All neurons gather information from the whole presented input.
Both populations consist of adaptive exponential integrate-and-fire neurons (AdEx) as described in Clopath et al. (2010) \cite{Clopath2010}.\\
The description of the membrane potential $u$ is presented in Eq.\ref{equ_Neuron}.
The slope factor is $\delta_T$, $C$ is the membrane capacitance, $E_L$ is the resting potential and $g_L$ is the leaking conductance.
The input is denoted by $I_{exc}$ for excitatory and $I_{inh}$ for inhibitory current.

\begin{equation}
C \frac{du}{dt} = -g_L(u-E_L)+g_L \Delta_T e^{ \frac{u-V_T}{\Delta_T}} - w_{ad}+z+I_{exc} - I_{inh}
\label{equ_Neuron}
\end{equation}

A spike is emitted, when the membrane potential exceeds the adaptive spiking threshold $V_T$.
After a spike, the membrane potential is set to $29mV$ for $2ms$, and then it is set back to $E_L$.
Furthermore, $V_T$ is set to $V_{T_{max}}$. The spiking threshold will decay towards to $V_{T_{rest}}$ with a time constant $\tau_{V_T}$.
\begin{equation}
\tau_{V_T} \frac{dV_T}{dt} = -(V_T - V_{T_{rest}})
\label{equ_VT}
\end{equation}
The depolarizing after potential is described by $z$.
After a spike, it is set to $I_{sp}$ and decays otherwise towards to zero with a time constant $\tau_z$.
The hyperpolarizing adaption current $w_{ad}$ is increased by an amount of $b$, when the neuron spikes.
Otherwise, it decays exponentially to the resting potential.
\begin{equation}\label{equ_Z}
\tau_z \frac{dz}{dt} = -z
\end{equation}

\begin{equation}
\tau_{w_{ad}}\frac{dw_{ad}}{dt} = a(u-E_L)-w_{ad}
\label{equ_Wad}
\end{equation}

The input currents are incremented by the sum of the presynaptic spikes of the previous time step, multiplied with the synaptic weight.
For excitatory synapses the current decays with a short time constant $\tau_{I_{exc}}$ of $1ms$. With $10ms$ as a time constant $\tau_{I_{inh}}$, the inhibitory synapses decay slower.
In Eq.\ref{equ_Curr} is the calculation of excitatory current exemplary, where $\sigma$ is the indicator function and $t'_i$ is the time point of a presynaptic spike.

\begin{equation}
\tau_{I_{exc}}\frac{dI_{exc}}{dt} = -I_{exc} + w^{exc}_i\sum_{i \in Exc}\delta(t-t_i^{'})
\label{equ_Curr}
\end{equation}
\subsubsection{Excitatory plasticity.}

The plasticity of the excitatory connections from the first to the second layer, as well as connections from the excitatory to the inhibitory population within the second layer, follows the voltage-based STDP rule \cite{Clopath2010}.
The development of the weight between a presynaptic neuron $i$ and a postsynaptic neuron depends on the presynaptic spike event $X_i$ and the presynaptic spike trace $\overline{x}_i$
as well as on the postsynaptic membrane potential $u$ and two averages of the membrane potential $\bar{u}_{+}$ and $\bar{u}_{-}$ as described in Eq.\ref{eqn_ubar} exemplary for $\bar{u_+}$ and a time constant $\tau_{+}$.
The definition of $\bar{u}_-$ is analog but with a different time constant $\tau_{-}$.

\begin{equation}\label{eqn_ubar}
\tau_+ \frac{d \bar{u}_+}{dt} = -\bar{u}_+ + u,
\end{equation}

The parameters $A_{LTP}$ and $A_{LTD}$ are the learning rates for long-term potentiation (LTP) and long-term depression (LTD).
Both parameters $\theta_{+}$ and $\theta_{-}$ are thresholds, which must be exceeded by the membrane potential or its long time averages.
\begin{equation}
\frac{dw_{i}}{dt} = A_{LTP} \ \overline{x}_i(u-\theta_+)^+ (\overline{u}_+ - \theta_-)^+ -  A_{LTD} \frac{\bar{\bar{u}}}{u_{ref}} X_i(\overline{u}_- - \theta_-)^+
\label{equ_STDP}
\end{equation}

The homoeostatic mechanism of the learning rule is implemented by the ratio between $\bar{\bar{u}}$ and the reference value $u_{ref}$ (Eq.\ref{equ_homoeo}).
It adjusts the amount of emergent LTD to control the postsynaptic firing rate.
Therefore,  $\bar{\bar{u}}$  implements a sliding threshold to develop selectivity of the neurons.

\begin{equation}
\tau_{\bar{\bar{u}}} \frac{d\bar{\bar{u}}}{dt} = [(u-E_L)^+]^2 - \bar{\bar{u}}
\label{equ_homoeo}
\end{equation}

Clopath et al. (2010) \cite{Clopath2010} propose to equalize the norm of the OFF weights to the norm of the ON weights every $20 s$.
We did this for the excitatory weights from the input layer to excitatory and the inhibitory population, per neuron.
The weights are limited by an upper and lower bound.

\subsubsection{Inhibitory plasticity.}

The connections from the inhibitory to the excitatory population and the lateral connections between the inhibitory neurons develop with the inhibitory learning rule from Vogels et al. (2011) \cite{Vogels2011} (see Eq.\ref{equ_iSTDP}).
\begin{eqnarray}
\Delta w_{ij} &= & \eta(\bar{x}_{j}-\rho) \;\; \textnormal{, for pre-synaptic spike } \\
\Delta w_{ij} &= & \eta(\bar{x}_{i}) \;\;\;\;\;\;\;\;\, \textnormal{, for post-synaptic spike} \nonumber
\label{equ_iSTDP}
\end{eqnarray}
The pre-synaptic spike trace is $\bar{x}_{i}$ and the spike trace for the post-synaptic neuron is $\bar{x}_j$.
When the particular neuron spikes, the spike trace increases with one, otherwise it decays with $\tau_{i}$ or $\tau_{j}$ to zero.
The inhibitory weight changes on a pre- or postsynaptic spike with the learning rate $\eta$.
A highly active post-synaptic neuron leads to a higher spiking trace $\bar{x}_{j}$ and to an increase of the inhibitory weights.
The constant value $\rho$ specifies the strength of inhibition to suppress the postsynaptic activity until LTD can occur.
The inhibitory weights are limited by a lower and upper bound.

\subsection{Deep Convolutional Networks}

To assess the performance of our network approach on MNIST recognition, we compared it to two deep convolutional neural networks (DCNN).
The first network is the well known LeNet 5, introduced from LeCun et al. (1998) \cite{LeCun1998}. 
It is hierarchically structured with two pairs of 2D-convolutional and max-pooling layers, followed by two fully connected one-dimensional layers.
The last layer is the classification layer with a \textit{"softmax"} classifier.
The first convolutional layer has a kernel size of $3\times3$ pixels and $32$ feature maps.
The kernel size of the second convolutional layer is $3\times3$ too, but consists of $64$ feature maps.
For the second max-pooling layer, a dropout regularisation with a dropout ratio of $0.5$ is used.
Both max-pooling layers have a $2\times2$ pooling size.\\
The architecture of the second model is based on the VGG16 network proposed by Simonyan and Zisserman (2015) \cite{Simonyan2015}.
As a consequence of the small size of the input, why we have to remove the last three 2D convolutional and the 2D max-pooling layer.
Further,  no dropout regularisation was	done.
This shortened model is further called $VGG13$.
Both networks are learned for $50$ epochs on the MNIST training set \cite{LeCun1998}.
The validation accuracy is measured on $10 \%$ of the training set.
The remaining $90 \%$ are used for learning.
The adadelta optimizer \cite{Zeiler2012} with $\rho=0.95$ is used for both networks.

\subsection{Measurement of accuracy}

The MNIST images have a resolution of $28\times28$ pixels. Because of the input size of the spiking network with $18\times18$ pixels, we divided each image of the MNIST set into four patches with each $18\times18$ pixel size. 
The first patch was cut out at the upper left corner and a horizontal and vertical pixel shift of 10 pixels was done to cut out the other three patches.
We presented every patch for $125ms$, without learning, and measured the number of spikes per neuron.
We repeated every patch presentation ten times to calculate a mean activity per neuron on every patch.
The activity vectors of the four patches corresponding to one digit are merged together.
For every digit, the final activity vector consists of $324 \times 4 = 1296$ values.
Further on, we fitted a support vector machine (SVM) with the merged activity vectors of the train set.
Before the fitting, we normed the activity vectors between zero and one. 
The SVM had a linear kernel, the squared hinge loss and the L2 penalty with a $C$-parameter of one.
To measure the accuracy, we used the merged activity vectors of the test set as input to the fitted SVM and compared the known labels with the predictions of the SVM.
Finally, we measured the accuracy of five separately learned networks and will present the average accuracy here.\\
We measured the accuracy of both DCNNs by presenting the MNIST test set and comparing their prediction with the known labels.
As for the spiking network, we measured the accuracy of five separately learned networks and present the average accuracy here.

\subsection{Robustness against pixel erasement}

\begin{figure}
	\centering
	\includegraphics[width=.55\textwidth]{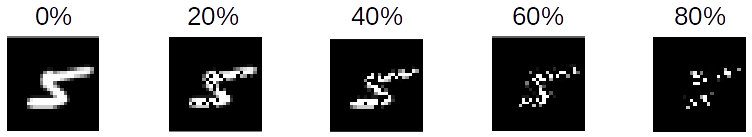}
	\caption{Example for the image distortion from the original digit to $80\%$ pixel erasement.}
	\label{fig_pE}
\end{figure}

In a previous study Kermani et al. (2015) demonstrated, that networks with biologically motivated learning rules in combination with inhibitory synapses are more robust against
the loss of information in the input.
They measured the classification accuracy of their network for different levels of pixel erasement in the MNIST dataset \cite{Kermani2015}.
Following this approach, we erased the pixels of all digits in the MNIST test sets in $5\%$ steps, erasing only pixels with a value above zero (see Fig.\ref{fig_pE}).
We created one data set per erasement level and showed each model the same dataset.
For each level of pixel erasement we measured the number of correct classifications as mentioned above.
Independently from the number of erased pixels, the SVM has always been fitted with the activity vectors measured on the original training set.

\section{Results}

\begin{figure}
	\centering
	\includegraphics[width=.49\textwidth]{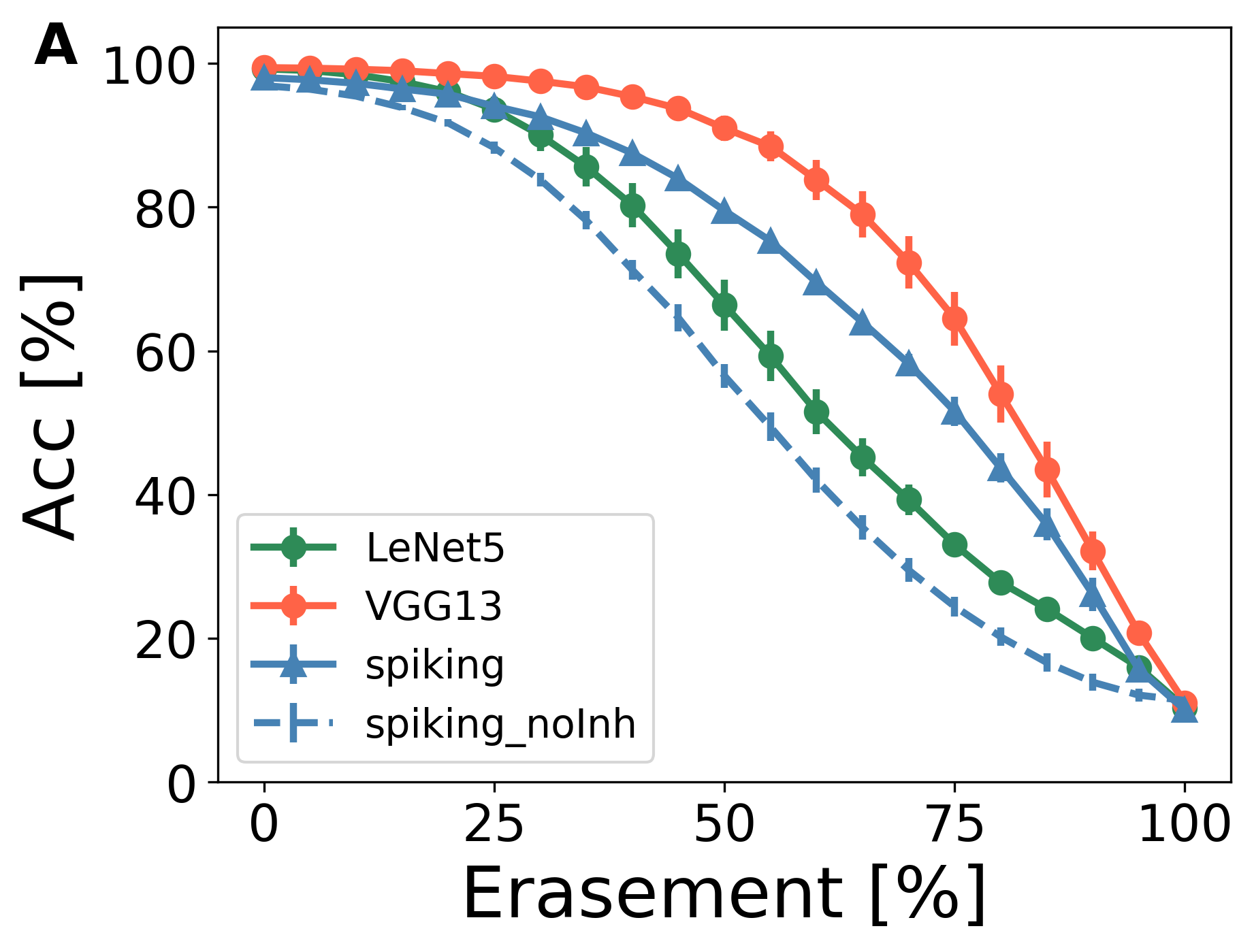}
	\includegraphics[width=.49\textwidth]{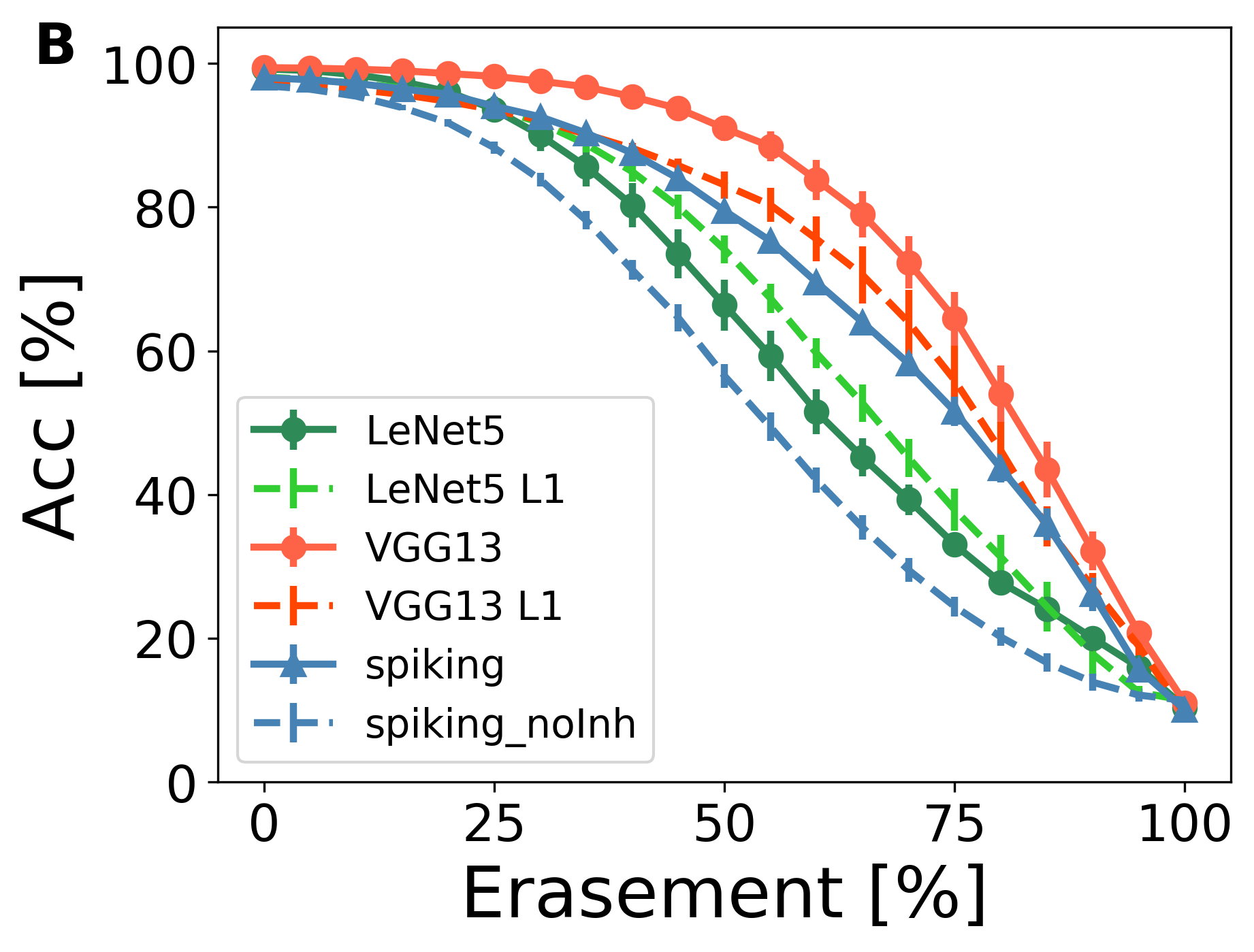}
	\caption{Classification accuracy as a function of the level of pixel erasement.
		\textbf{A}, Robustness of the spiking network (blue line) is between the LeNet 5 (green line) and VGG13 (red line) network. Deactivation of inhibition leads to a less robust spiking network (dashed blue line).
		\textbf{B}, The first layer of LeNet 5 (dashed green line) is more robust than the complete network. Whereas the first layer of the VGG13 is less robust (dashed red line).}
	\label{fig_ACC}
\end{figure}

Our network achieved on the original MNIST test data set an average accuracy of $98.08 \%$ over five runs. If the inhibition is removed, $96.81 \%$ accuracy is archived.
The LeNet 5 implementation achieved $99.24 \%$ and the
VGG13 network $99.41\%$, averaged over five runs (Tab.\ref{tab_DCN}).
Our results show, that at $25\%$ erased pixels the spiking network achieves higher accuracy values than the LeNet 5 network, but lower values than the VGG13 network.
By setting all inhibitory synapses to zero on the finished trained spiking network, we deactivated the inhibition and measured again the accuracy on the different levels of pixel erasement.
Doing that, we have been able to reproduce the result from Kermani et al. (2015) \cite{Kermani2015}.
As mentioned in the original publication, the accuracy decreases without inhibition stronger than with it (see Fig.\ref{fig_ACC} \textbf{A}). \\
In contrast to the multi layer deep convolutional networks the spiking network only consists of one layer of excitatory neurons.
Because of that, we measured the accuracy of the LeNet 5 and VGG13 only with the activity of the first convolutional layer.
Therefore, the output of the first layer was connected to a classification layer with $10$ units and a softmax activation function.
Only the weights from the convolutional to the classification layer were trained on the MNIST training set.
The classification on the pixel erased dataset was done as for the other deep networks.
With an accuracy of $98.1\%$ from the first layer of LeNet 5 and $97.29\%$ of the first layer of the VGG13, the first convolutional layer alone achieved a lower accuracy on the original MNIST test set than the complete network  (Tab.\ref{tab_DCN}).
By stepwise pixel erasement, the first layer of the LeNet 5 is slightly robuster than the complete network.
In contrast the first layer of the VGG13 model is less robust than the complete model.
The course of the curve is similar to the spiking network (Fig.\ref{fig_ACC} \textbf{B}).

\begin{figure}
	\centering
	\includegraphics[width=.49\textwidth]{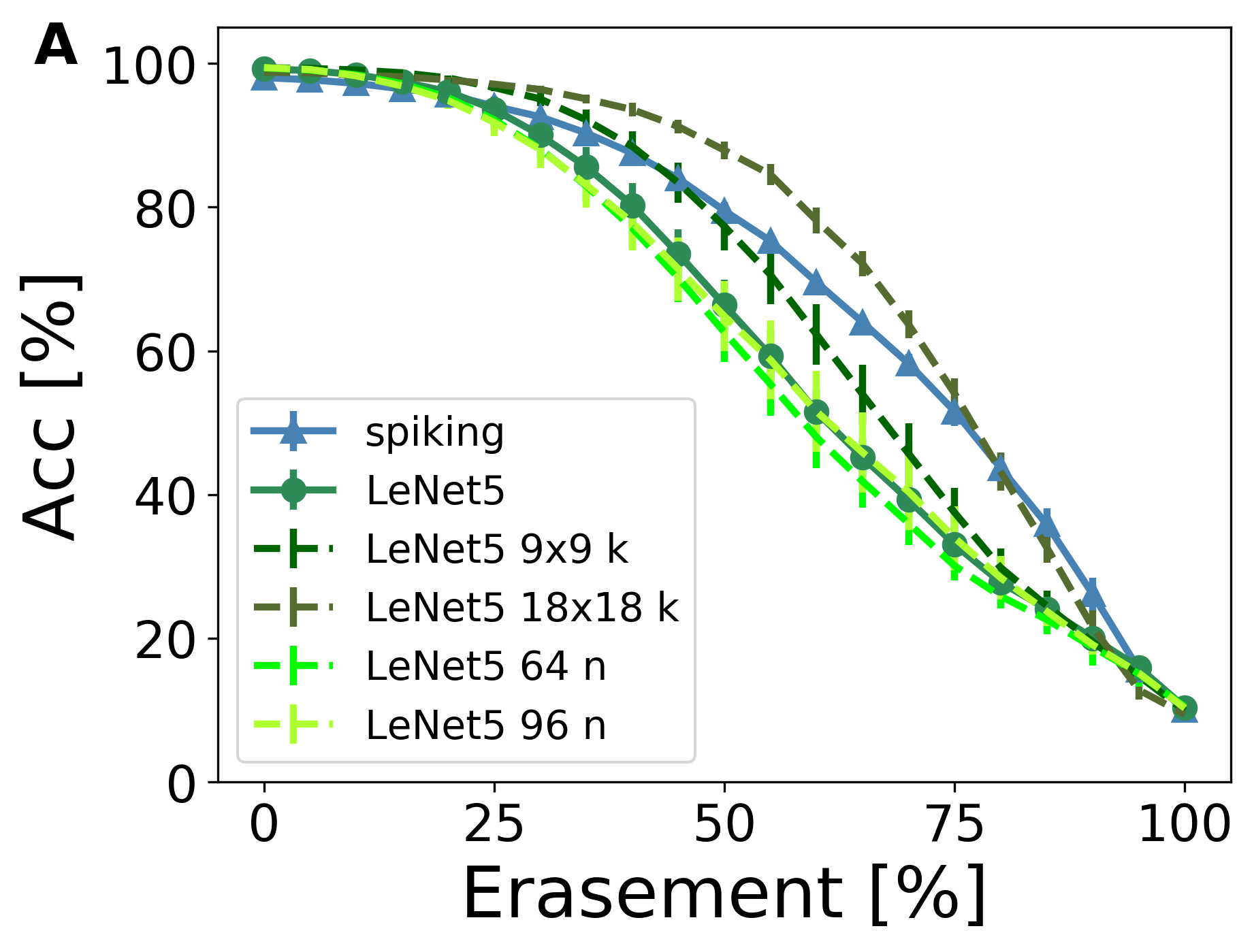}
	\includegraphics[width=.49\textwidth]{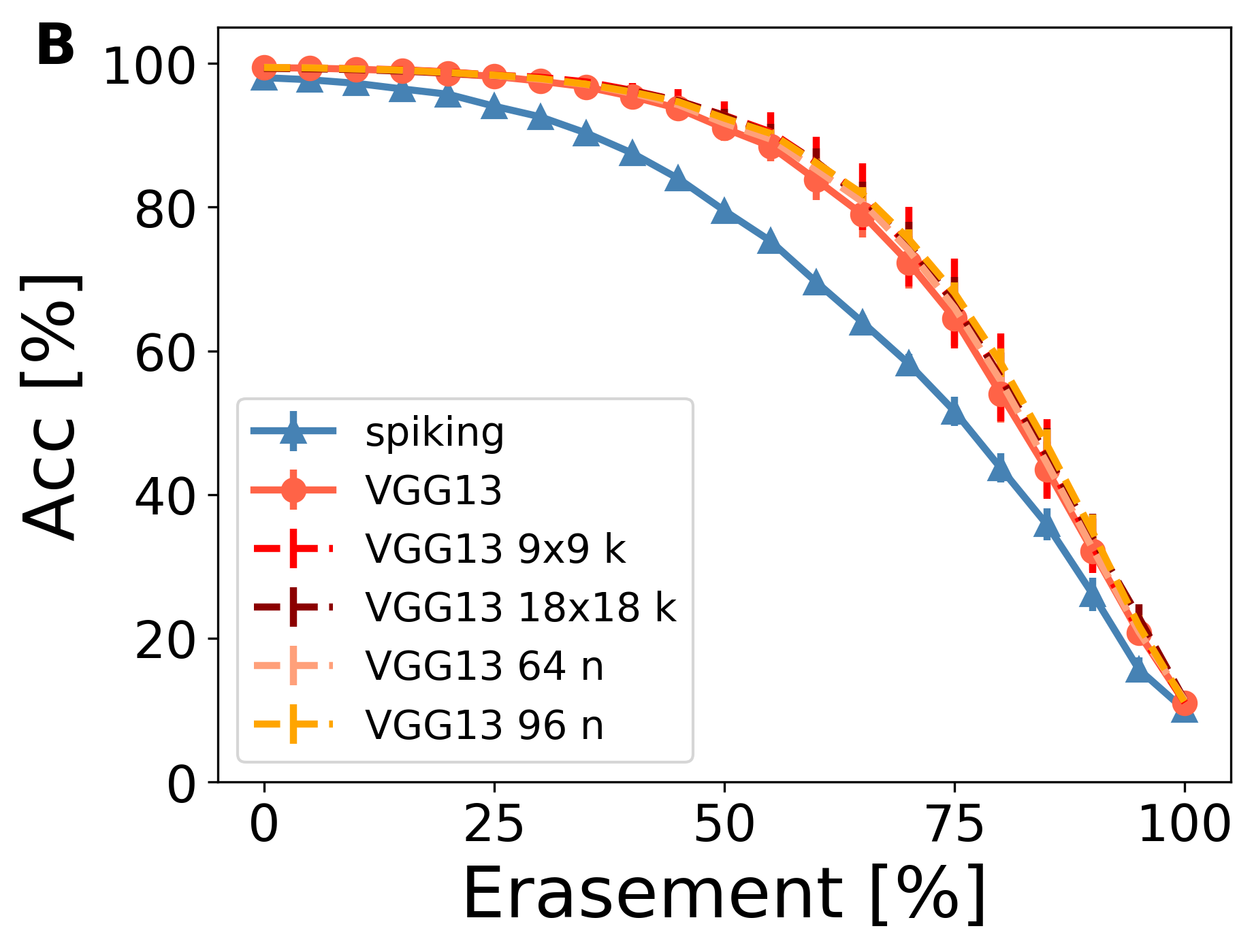}
	\caption{Classification accuracy for different configurations of the DCNNs.
		\textbf{A} LeNet 5 with different numbers of feature maps (brighter lines) and larger kernel sizes (darker lines).
		\textbf{B} VGG13 with different numbers of feature maps (brighter lines) and larger kernel sizes (darker lines).
		More feature maps shows no change or slightly less robustness against pixel erasement. Larger kernel sizes lead to an improved robustness in LeNet5.}
	\label{fig_ACC2}
\end{figure}

The size of the receptive fields in the spiking model does not correspond to the size of the convolutional kernel in the DCNNs.
Further, every feature map in the convolutional layer shares the same convolutional kernel.
Our spiking network learns 324 different receptive fields.
That would be equivalent to 324 different feature maps in a DCNN.
To accommodate these differences between the spiking approach and the DCNNs, we changed the number of feature maps in the first convolutional layer and the kernel size in the LeNet 5 and the VGG13 network to $9\times9$ and $18\times18$.
To avoid unnecessary computational load and the possibility of over fitting
we increased the number of feature maps only to $64$ and $96$.
The increased kernel size in the LeNet 5 implementation leads to a significant improvement of the robustness (see Fig.\ref{fig_ACC2} \textbf{A}).
However, for the VGG13 model, it does not lead to a significant change (see Fig.\ref{fig_ACC2} \textbf{B}).
An increased number of feature maps in both DCNNs seems to have no effect on the robustness.

\begin{table}
	\begin{tabular}{l|c|c|c|c|c|c}
		architecture & normal & first layer only & 64 features & 96 features & 9x9 kernel & 18x18 kernel \\
		\hline
		\hline
		LeNet 5		 &	$99.24\%$ &	$98.10\%$	 &	$99.38\%$  & $99.42\%$	&	$99.03\%$  &	$98.77\%$ \\
		\hline
		VGG13		 &	$99.41\%$ &	$97.29\%$	 &	$99.44\%$  & $99.43\%$	&	$99.41\%$ &	$99.32\%$ \\
	\end{tabular}
	\caption{Accuracy values on the deep convolutional networks LeNet 5 and VGG13, with different number of features and sizes for the kernel filter. Measured on the original MNIST test set. Averaged over five runs per model.}
	\label{tab_DCN}
\end{table}

\section{Discussion}

Our proposed two layer spiking neural network (SNN) archived an accuracy of $98.08 \%$ on the original MNIST data set.
Previous unsupervised learned SNN have shown slightly weaker results on the MNIST data set \cite{Diehl2015,Kheradpisheh2017}.
Diehl and Cook (2015) \cite{Diehl2015} presented a two layer SNN with a similar architecture to the here proposed one.
They achieved an accuracy of $95.0\%$ with 6400 neurons and an accuracy of $87.0\%$ with 400 neurons.
In contrast to our spiking network, the excitatory population in their network is one-to-one connected to the inhibitory one.
A spike of an excitatory neuron leads to a spike of the corresponding inhibitory neuron.
This sends an inhibitory signal back in the excitatory population to implement lateral inhibition.
Second, each neuron was connected to the full input of the MNIST data set and thus learned complete digits as receptive fields.
After learning, they assigned every neuron a class, referred to the class with the highest activity on the training set \cite{Diehl2015}.
The class of the most active neuron defined the prediction of the network on the test set.
Our network is learned on natural scene input \cite{Olshausen1996} instead of images of the MNIST data set.
Because presenting each neuron just a small segment of different spatial orientations and spatial frequencies, our network learns Gabor-like receptive fields \cite{Clopath2010}.
These feature detectors are selective for only a part of the presented input instead of a complete digit.
Further on, the classification for our approach is done by training a simple linear SVM with activity vectors of the excitatory population.
Instead of only considering the activity of the most active neuron, here the classification includes the activity of all excitatory neurons.
The neurons are active on a part of the digit, defined by the receptive field.
Therefore, different digits are decoded by the combination of different neuronal activities.
This leads to a better classification accuracy with a smaller number of neurons.\\
Another unsupervised spiking network was presented by Tavanaei and Maida \cite{Tavanaei2017}.
In contrast to our network, their network consists of four layers.
Further, the input consists of $5\times5$ pixels sized overlapping patches, cut out of the MNIST training set.
Every pixel value determines the rate of the input spike train for the neurons in the second layer.
In the second layer exists lateral inhibitory connections between the neurons.
This lead to Gabor-like receptive fields in the second layer.
The next layer was a max-pooling layer, followed by a so called 'feature discovery' layer.
After learning in the second layer was finished, they learned the fourth layer.
The output of the fourth layer was used to train a SVM for the classification. 
They used four SVMs with different kernels and averaged them.
With $32$ neurons in the second and $128$ neurons in the last layer they archived an accuracy of $98.36\%$ on the MNIST test.\\
A deeper unsupervised spiking approach was presented by Kheradpisheh et al. (2017) \cite{Kheradpisheh2017}.
They presented a deep spiking network to mimic convolutional and max-pooling layers by using a temporal coding STDP learning algorithm.
This means, that the first firing neuron learned the most, while later firing neurons learned less or nothing.
Their network consists of three pairs of a convolutional and a max-pooling layer.
For the classification, they used a linear SVM on the output of the last pooling layer.
On the MNIST data set, they achieved an accuracy of $98.4 \%$ \cite{Kheradpisheh2017}.
However, the used temporal coding implements a "winner takes it all" mechanism.
This is less biologically plausible than the used learning rules in our approach. 
Nonetheless, the complex structure of the network from Tavanaei and Maida (2017) \cite{Tavanaei2017} and of the Kheradpisheh et al. (2017) \cite{Kheradpisheh2017} network is an evidence for the possibility of unsupervised STDP learning rules in a multi-layer network.\\
A comparison with two deep convolutional networks on stepwise pixel erasement showed, that our LeNet 5 implementation is less robust and the VGG13 model is more robust than the here proposed spiking network (Fig.\ref{fig_ACC} \textbf{A}).
In case of the accuracy only been measured on the activity of the first layer, the LeNet 5 first layer is more robust than the complete model. In the VGG13 the first layer is less robust.
The first convolutional layer of both models has the same kernel size ($3\times3$) and number of features ($32$), but the robustness of both layers is different (Fig.\ref{fig_ACC} \textbf{B}).
Both deep convolutional neural networks (DCNNs) have different numbers of layers and a different order of convolutional and max-pooling layers.
This suggests, that the structure of the network influences the learning result in the first convolutional layer, especially how the error between output and input is back propagated.
In contrast to an increase of the number of features, an increase of the convolutional kernel size leads to an improvement of the robustness(Fig.\ref{fig_ACC2}), but to a decrease in the accuracy on the original data set by the LeNet 5 model (Tab.\ref{tab_DCN}).
An increase of the number of features or the convolutional kernel size does not lead to a significant change for the VGG13 model.
With a larger filter kernel, the erasement of a fixed number of pixels in the input has a lower influence on the activity of the neurons.
In a $3\times3$ kernel three erased pixels in the input cause a loss of $33.33 \%$ of the incoming activity.
In a $9\times9$ kernel the loss is only $3.7 \%$.\\
As mentioned in previous works \cite{Kermani2015}, our results show that learned lateral inhibition leads to an improvement of the classification robustness against pixel erasement in unsupervised learned models.
On one side, neurons loose the sharpening of their selectivity without inhibition \cite{Priebe2008,Katzner2011}.
On the other side, the correlation between the neuron activities increases. This leads to less distinct input encoding, that in turn decreases the robustness against pixel erasement \cite{Kermani2015}.
The robustness in DCNNs is influenced by the learned feature maps as a result of the back propagation mechanism and the network architecture.
Further, a larger size of the kernel filter improves the robustness. Whereas the number of feature maps are not that relevant.
The absence of inhibition in DCNNs suggest, that not only the influence of inhibition on the neuronal activity improves the robustness.
Rather the filter size and the structure of the learned filters are important for a robust behaviour.

\subsection*{Acknowledgement}

This work was supported by the European Social Fund (ESF) and the Freistaat Sachsen.

\end{document}